%% file: main.tex
\documentclass[10pt,twocolumn,letterpaper]{article}

\usepackage{wacv}              
\usepackage{tikz}
\usepackage[accsupp]{axessibility}  
\usetikzlibrary{positioning,arrows.meta}
\usepackage{tabularx}
\usepackage{booktabs} 
\usepackage{subcaption} 
\input{preamble}

\definecolor{wacvblue}{rgb}{0.21,0.49,0.74}
\usepackage[pagebackref,breaklinks,colorlinks,allcolors=wacvblue]{hyperref}

\def\wacvPaperID{15} 
\def\confName{WACV}
\def\confYear{2026}

\title{TFFM: Topology-Aware Feature Fusion Module via Latent Graph Reasoning for Retinal Vessel Segmentation}

\author{
Iftekhar Ahmed$^\ast$ \quad 
Shakib Absar$^\ast$ \quad 
Aftar Ahmad Sami$^\dagger$ \quad 
Shadman Sakib$^\S$ \\
Debojyoti Biswas$^\ddagger$ \quad
Seraj Al Mahmud Mostafa$^\S$ \\[0.5em]
$^\ast$Leading University, Sylhet, Bangladesh, 
$^\dagger$University of Houston, Houston, TX, USA \\
$^\S$University of Maryland, Baltimore County, Baltimore, MD, USA \\
$^\ddagger$Pennsylvania State University, University Park, PA, USA \\[0.3em]
{\small \texttt{$^\ast$\{iftekharifat007, sabsar42\}@gmail.com}}, 
{\small \texttt{$^\dagger$asami5@uh.edu}} \\
{\small \texttt{$^\S$\{ssakib1, serajmostafa\}@umbc.edu}}, 
{\small \texttt{$^\ddagger$dbb5917@psu.edu}}
}

\begin{document}
\maketitle

\begin{abstract}
Precise segmentation of retinal arteries and veins carries the diagnosis of systemic cardiovascular conditions. However, standard convolutional architectures often yield topologically disjointed segmentations, characterized by gaps and discontinuities that render reliable graph-based clinical analysis impossible despite high pixel-level accuracy. To address this, we introduce a topology-aware framework engineered to maintain vascular connectivity. Our architecture fuses a Topological Feature Fusion Module (TFFM) that maps local feature representations into a latent graph space, deploying Graph Attention Networks to capture global structural dependencies often missed by fixed receptive fields. Furthermore, we drive the learning process with a hybrid objective function, coupling Tversky loss for class imbalance with soft clDice loss to explicitly penalize topological disconnects. Evaluation on the Fundus-AVSeg dataset reveals state-of-the-art performance, achieving a combined Dice score of 90.97\% and a 95\% Hausdorff Distance of 3.50 pixels. Notably, our method decreases vessel fragmentation by approximately 38\% relative to baselines, yielding topologically coherent vascular trees viable for automated biomarker quantification. We open-source our code at \url{https://tffm-module.github.io/}.
\end{abstract}

\section{Introduction}
\label{sec:introduction}

The human retina offers a unique, non-invasive window into the systemic microvasculature, where the geometric properties of arteries and veins ($A/V$) serve as potent biomarkers for cardiovascular disease, hypertension, and diabetic retinopathy. Currently, the standard of care relies on qualitative visual inspection or manual tracing, which is time-consuming, subjective, and prone to inter-observer variability \cite{khandouzi2022retinal}. However, for these biomarkers to be clinically actionable, automated segmentation must yield more than just pixel-level accuracy; it must preserve the topological integrity of the vascular graph. To a clinician, a blood vessel is a continuous transport network; to a standard Convolutional Neural Network (CNN), it is merely a collection of disjoint pixels. This semantic gap is the primary cause of the shattered vessel phenomenon, where deep learning (DL) models achieve high Dice scores yet produce fragmented, topologically invalid segmentations that render graph-based analysis impossible.

In this work, we propose a topology-aware deep learning framework especially designed to address the dual challenges of class imbalance and structural continuity in $A/V$ segmentation. Our approach is the result of a systematic ablation study identifying the optimal selection between architecture, encoder capacity, and loss formulation. In summary, we present a topology-aware framework that captures capillary regions. To overcome this, we introduce two key innovations. \textbf{\textit{1) Topological Feature Fusion Module (TFFM):}} We propose a novel module that projects feature maps into a latent graph space. By utilizing Graph Attention Networks (GATs) \cite{gat} within the decoder, the TFFM models the global connectivity of the vascular tree. It allows the network to reason about vessel continuity beyond the local receptive field of standard convolutions, and \textbf{\textit{2) Hybrid Topology-Aware Objective:}} We use a composite loss function combining Tversky loss \cite{salehi2017tversky}, which addresses the foreground-background imbalance, with soft clDice loss \cite{shit2021cldice}. The latter acts as a topological regularizer, penalizing breaks in the vessel skeleton and ensuring the predicted morphology aligns with the curvilinear nature of the vasculature.

\section{Related Works}
\label{sec:related_work}
Retinal vessel segmentation has evolved toward deep learning architectures, achieving high pixel-level accuracy, yet clinical utility depends on preserving vascular topology for biomarker extraction and graph-based analysis. Standard metrics (Dice, IoU) measure only region overlap, permitting topologically invalid segmentations where vessels fragment into disconnected components.

\textbf{Multi-Scale Feature Architectures.}
Liu et al.~\cite{liu2024imff} introduced IMFF-Net, achieving 0.9621 accuracy on DRIVE through Attention Pooling and multi-level feature fusion. However, optimizing solely with Dice loss yields no topology guarantees: high accuracy can mask severe fragmentation. Wang et al.~\cite{wang2024multiattention} proposed Multi-Scale Attention Fusion to preserve spatial information during pooling. While capturing fine vessels, their local spatial attention cannot enforce global connectivity, resulting in broken bifurcations despite strong pixel metrics. Zhou et al.~\cite{zhou2024multiscale} combined dilated residuals with dual attention, but their hybrid loss remains topology-agnostic. These methods exemplify a critical limitation: architectural improvements for feature extraction do not translate to topological validity without explicit connectivity modeling.

\textbf{State Space Models for Vessel Segmentation.}
Mamba-based architectures offer efficient long-range modeling. Wang et al.~\cite{wang2025hmmamba} proposed HM-Mamba with tubular structure-aware convolution, achieving $83.27\%$ Dice on DRIVE. However, HM-Mamba lacks topology metrics, failing to distinguish continuous trees from fragmented predictions. Similarly, attention-enhanced Mamba~\cite{hu2025multiscale} and TA-Mamba~\cite{shao2025tubular} improve feature modeling but lack graph-based reasoning about connectivity. Their evaluations emphasize sensitivity and specificity but omit skeleton-based metrics (clDice, junction detection) essential for structural fidelity. The fundamental issue persists: sequence modeling captures local dependencies but not global topological constraints.

\textbf{Topology-Preserving Loss Functions.}
Shit et al.~\cite{shit2021cldice} introduced clDice loss, computing skeleton similarity to enforce connectivity. While pioneering topology-aware optimization, clDice alone sacrifices pixel accuracy and suffers from noise sensitivity. Zhang et al.~\cite{zhang2024clce} proposed clCE loss, improving robustness but remaining purely loss-based. Yeganeh et al.~\cite{yeganeh2023scope} developed SCOPE, encoding continuity as training constraints. These approaches share a critical weakness: applying topology constraints only during optimization, not feature extraction. The network architecture lacks components explicitly designed to model connectivity, limiting propagation of topological information through the feature hierarchy. Loss functions guide learning, but cannot compensate for architectures incapable of reasoning about vessel relationships.

\textbf{Graph Neural Networks and Transformers.}
Li et al.~\cite{li2022dualdcgcn} proposed DE-DCGCN-EE with dynamic-channel graph convolution. However, their graphs operate on channel dimensions rather than spatial topology, missing geometric vessel relationships. Jalali et al.~\cite{jalali2024vganet} developed VGA-Net with pixel-level graph nodes, but this design is prohibitively expensive at high resolutions. Wang et al.~\cite{wang2022danet} combined transformers with adaptive upsampling (MICCAI 2022), while Tong et al.~\cite{TONG2024213} introduced a lightweight LiViT-Net. Despite achieving high accuracy, transformer methods report no connectivity metrics and face quadratic complexity constraints at medical imaging resolutions. Taufik et al.~\cite{taufik2025visual} demonstrated transformers' superiority for medical classification, yet their focus on interpretability leaves topology-preserving feature learning for segmentation unexplored.

\textbf{Towards Architecture-Integrated Topology Preservation.} Current methods optimize region overlap, often producing disconnected components despite high accuracy~\cite{liu2024imff}. Such fragmentation renders predictions clinically unusable, as standard metrics (Dice, IoU) ignore topological relationships~\cite{shit2021cldice,mosinska2018beyond}. A prediction achieving 95\% Dice but containing 50 disconnected fragments scores identically to a topologically correct segmentation under standard metrics, despite being fundamentally incompatible with downstream clinical analysis. The absence of topology-specific metrics (clDice~\cite{shit2021cldice}, Betti error, skeleton recall, junction detection) in most evaluations perpetuates this disconnect~\cite{mosinska2018beyond,yang2023directional}. To mitigate this issue, architectures can incorporate topological reasoning directly into feature extraction~\cite{yeganeh2023scope,zhang2024clce,li2020iternet}, enabling networks to learn representations that capture not only local vessel appearance but also the global connectivity constraints inherent to vascular trees.

In contrast to existing work, our method integrates topology preservation at both the architectural and loss levels through a TFFM with Graph Attention Networks and a hybrid Tversky + soft clDice objective that penalizes disconnections. This joint design produces topologically coherent vessel trees with substantially reduced fragmentation while preserving strong region-overlap performance suitable for downstream graph-based clinical analysis.

\section{Methodology }
\label{sec:methods}

\begin{figure*}[t]
    \centering
    \includegraphics[width=0.9\linewidth]{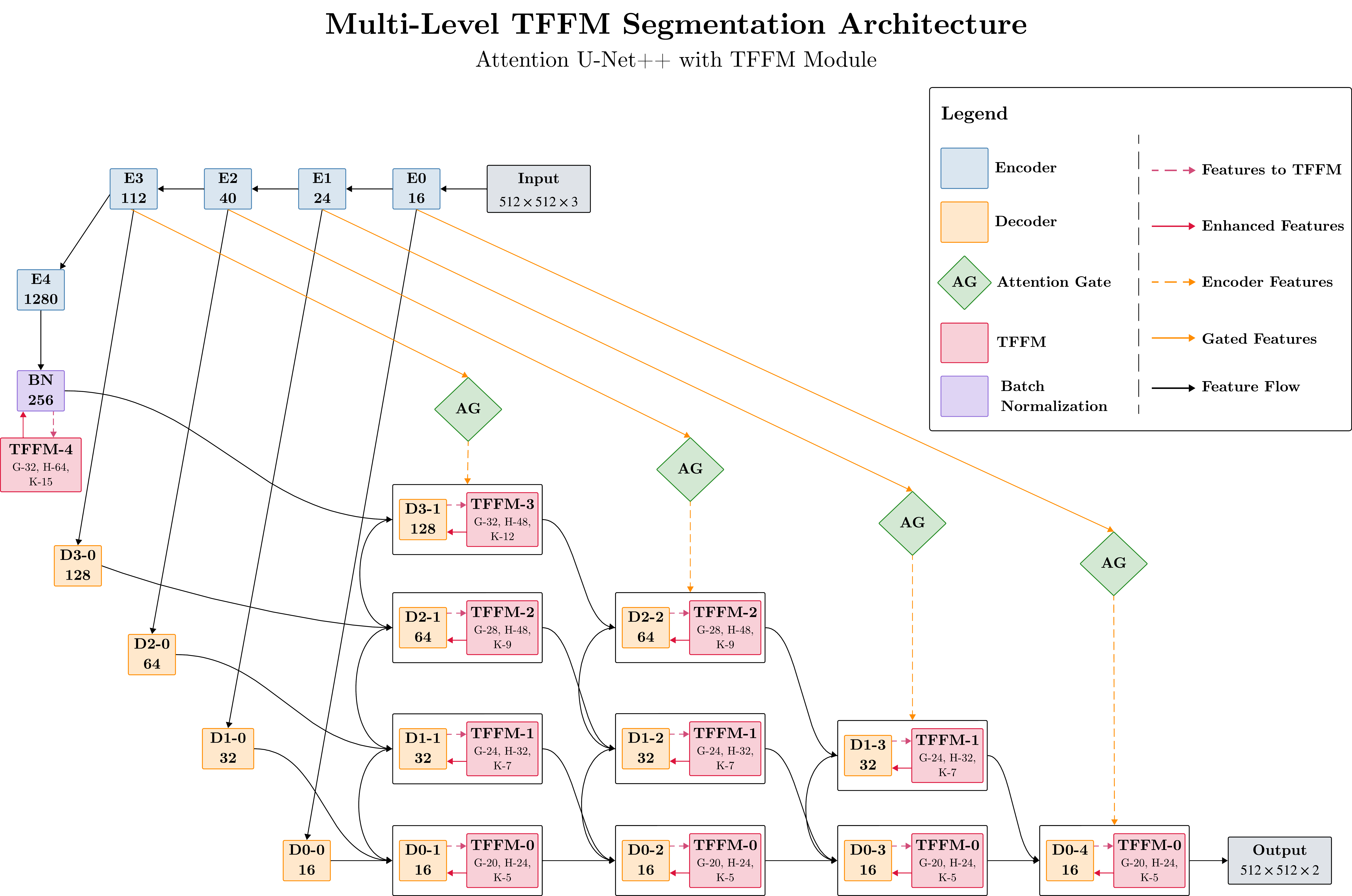}
    \caption{Overview of the proposed multi-level TFFM segmentation architecture.}
    \label{fig:architecture}
\end{figure*}

In this work, we propose a topology-aware segmentation framework tailored for retinal artery–vein delineation (Fig. \ref{fig:architecture}). The model builds upon a baseline identified through systematic ablation and is composed of a U-Net++ backbone enhanced with Attention Gates and an EfficientNet-B0 encoder. To preserve vascular connectivity, particularly in thin, tortuous structures, we introduce a Topological Feature Fusion Module (TFFM). TFFM explicitly captures and propagates topological cues across network levels. The training objective combines Tversky loss to mitigate class imbalance with soft clDice loss to enforce structural continuity and reduce topological errors.

\subsection{Baseline Architecture}
\label{sec:architecture}

The backbone architecture, established through comparative evaluation against U-Net \cite{ronneberger2015u}, Attention U-Net \cite{schlemper2019attention}, SegResNet \cite{myronenko20183d}, SwinUNETR \cite{hatamizadeh2021swin}, TransUNet \cite{chen2024transunet}, DeepLabV3+ \cite{chen2018encoder}, and Segformer \cite{xie2021segformer}, consists of U-Net++ \cite{zhou2018unet++} augmented with Attention Gates (AGs) \cite{schlemper2019attention}.

It has been selected for its optimal performance in capturing fine-grained morphological details compared to standard U-Net and U-Net variants during our preliminary experimentation. The complete ablation study supporting this choice is presented in Section~\ref{subsec:baseline_selection}.

For the encoder path, we substitute standard convolutional blocks with an EfficientNet-B0 backbone \cite{tan2019efficientnet}. Despite being a lighter model, EfficientNet-B0 achieved superior performance compared to the stronger EfficientNet-B4 variant in our experiments. This trend is affirmed by the full encoder ablation provided in Section~\ref{subsec:encoder_selection}.

\subsection{Topological Feature Fusion Module (TFFM)}
\label{subsec:tffm}

\begin{figure}[t]
    \centering
    \includegraphics[width=1\linewidth]{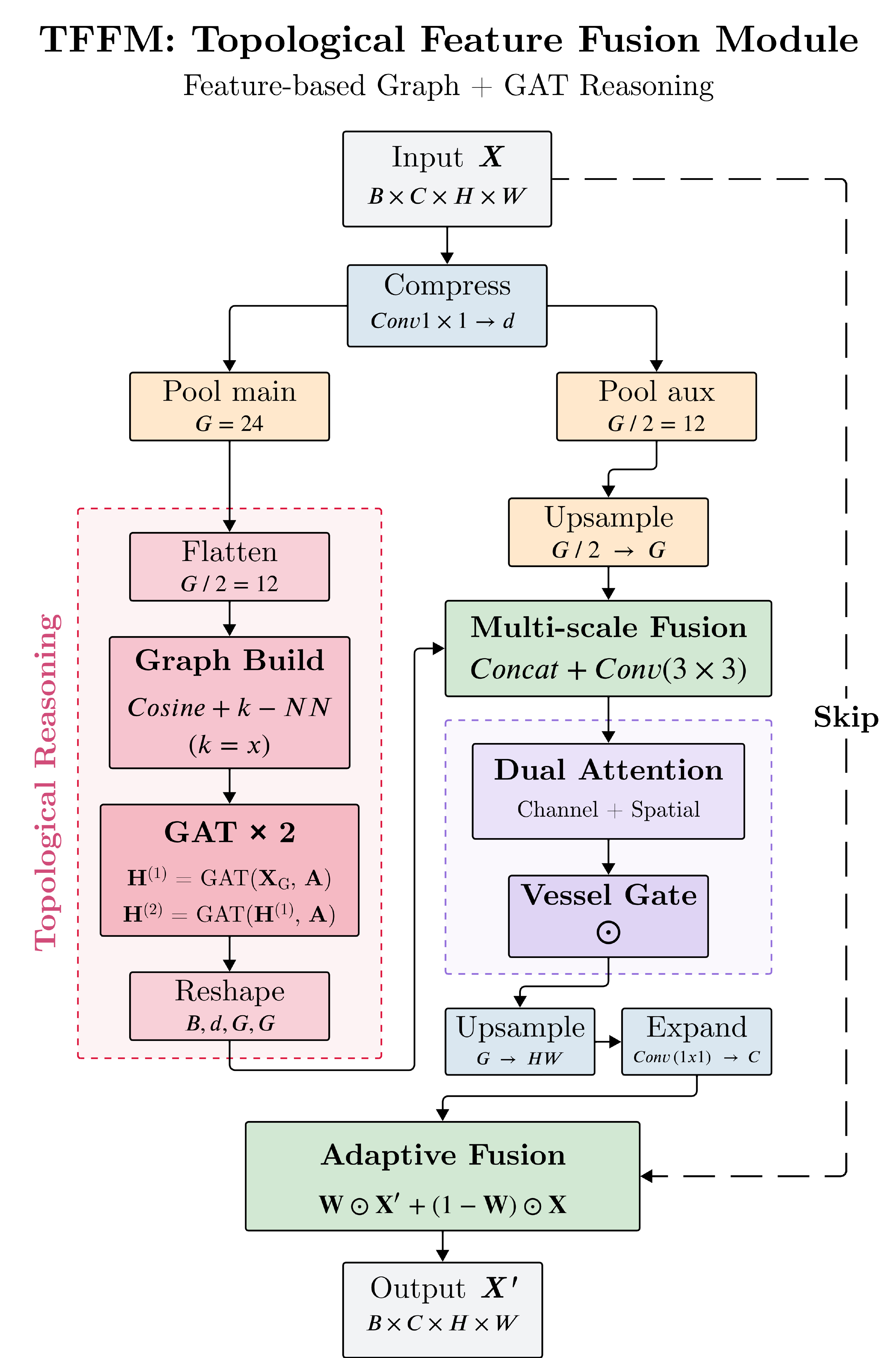}
    \caption{Overview of the proposed TFFM}
    \label{fig:tffm}
\end{figure}

To address the inherent limitations in structural fidelity, we introduce the TFFM. While the backbone captures local features, TFFM projects these features into a latent graph space to reason about global connectivity. The module (Figure \ref{fig:tffm}) operates in four stages: multi-scale abstraction, dynamic graph reasoning, attention-based refinement, and adaptive integration.
\vspace{-1.15em}

\paragraph{Scale-Adaptive Abstraction and Graph Construction:}
The TFFM is applied at each decoder level $l$ with depth-dependent hyperparameters: grid size $G_l$, hidden dimension $C_{h,l}$, and neighbor count $k_l$. The input $\mathbf{X}$ is first compressed to $C_{h,l}$ channels using a $1 \times 1$ convolution. Two pooled representations are then generated: a primary grid $\mathbf{F}_{main} \in \mathbb{R}^{C_{h,l} \times G_l \times G_l}$ and an auxiliary grid $\mathbf{F}_{aux}$ of size $\max(G_l/2, 4)$. This ensures a minimum coarse resolution.

We form a feature-space graph $\mathcal{G}=(\mathcal{V},\mathcal{E})$ where nodes are vectors in the flattened $\mathbf{F}_{main}$. A cosine similarity matrix $\mathbf{S}$ is computed and sparsified via top-$k_l$ selection:
\begin{equation}
    \mathbf{A}_{ij} = \mathbb{I}(j \in \text{top-}k_l(\mathbf{S}_{i:})), \qquad 
    \tilde{\mathbf{A}} = \mathbf{A} + \mathbf{I},
\vspace{-.8em}    
\end{equation}

\paragraph{Graph Attention:}
We utilize a Graph Attention Network (GAT) \cite{gat} with masked attention. The unnormalized attention scores are computed and masked by the adjacency structure:
\begin{equation}
    e_{ij} = \text{LeakyReLU}\!\left(\mathbf{a}^T[\mathbf{W}\mathbf{h}_i \| \mathbf{W}\mathbf{h}_j]\right)\cdot \tilde{\mathbf{A}}_{ij},
\end{equation}

\begin{equation}
    \alpha_{ij} = \text{softmax}_j (e_{ij} / \tau),
\end{equation}
where $\tau$ is a learnable temperature parameter. Non-neighbors (where $\tilde{\mathbf{A}}_{ij}=0$) contribute zero to the attention logits, effectively being pruned from the aggregation.

\paragraph{Multi-scale Fusion and Refinement:}
The output graph features are reshaped to $\mathbb{R}^{C_{h,l} \times G_l \times G_l}$ and concatenated with the upsampled auxiliary features. A $3\times3$ convolution produces the fused representation $\mathbf{F}_{fused}$. Channel Attention Module (CAM) and Spatial Attention Module (SAM) refine this tensor:
\begin{equation}
    \mathbf{F}_{att} = 
    \mathbf{F}_{fused} \odot 
    \text{CAM}(\mathbf{F}_{fused}) \odot 
    \text{SAM}(\mathbf{F}_{fused}),
\end{equation}

followed by a vesselness gating mechanism where $\mathcal{F}_{vessel}$ is a lightweight CNN that predicts vessel-relevant weights:
\begin{equation}
    \mathbf{F}_{ref} = \mathbf{F}_{att} \odot \sigma(\mathcal{F}_{vessel}(\mathbf{F}_{att})),
\end{equation}

\paragraph{Adaptive Integration:}
The refined features are upsampled to $(H,W)$ and expanded back to the original channel dimension. A learnable gated residual connection then blends local and topological information:
\begin{equation}
    \lambda = \sigma(\Psi_{gate}([\mathbf{X} \| \mathbf{F}_{exp}])),
\end{equation}
\begin{equation}
    \mathbf{Y} = 
    \lambda \odot \Psi_{fusion}([\mathbf{X} \| \mathbf{F}_{exp}]) 
    + (1-\lambda)\odot \mathbf{X},
\end{equation}
This gate allows the network to selectively inject global connectivity cues while preserving fine local structure.

\subsection{Loss Function}
\label{subsec:loss}

To address retinal artery/vein segmentation, we employ a hybrid loss function combining Tversky Loss ($L_{Tv}$) for handling the extreme foreground-background class imbalance and soft clDice Loss ($L_{clDice}$) for preserving topological connectivity of thin vessels. The total loss is averaged over both vessel classes.

\textbf{Tversky Loss:} We employ Tversky loss to mitigate foreground-background imbalance by asymmetrically weighting false positives and negatives. For predictions $P_c$ and ground truth $G_c$, we define the true positives $TP = \sum p_{ci}g_{ci}$, false negatives $FN = \sum (1-p_{ci})g_{ci}$, and false positives $FP = \sum p_{ci}(1-g_{ci})$. The Tversky index is formulated as:

\begin{equation}
TI(P_c, G_c) = \frac{TP}{TP + \alpha FN + \beta FP},
\end{equation}

In our experiments, we observed that missed vessel segments (false negatives) break the vascular topology more severely than background noise (false positives). Consequently, we tune the hyperparameters to prioritize recall. We set $\alpha = 0.65$ to increase the penalty for false negatives. Following the standard implementation in the MONAI framework, $\beta$ is set to $1 - \alpha = 0.35$. The Tversky loss is then computed as:

\begin{equation}
L_{Tv} = 1 - \frac{1}{C} \sum_{c=1}^{C} TI(P_c, G_c),
\end{equation}

\textbf{Soft clDice Loss: }While $L_{Tv}$ improves pixel-level accuracy, it does not explicitly penalize topological breaks. We therefore incorporate the soft clDice loss, which operates on the continuous probability maps to preserve connectivity. We apply a differentiable \textit{soft-skeletonization} function $S(\cdot)$ to both the predicted probabilities $P$ and the ground truth $G$. This algorithm uses iterative min/max-pooling operations to approximate the morphological skeleton on soft inputs:

\begin{equation}
\begin{aligned}
S(P) &= \text{SoftSkeletonize}(P), \\
S(G) &= \text{SoftSkeletonize}(G),
\end{aligned}
\end{equation}

Using these soft skeletons, we compute the Topological Precision ($T_{prec}$) and Topological Sensitivity ($T_{sens}$), which measure the overlap between the predicted soft skeleton and the ground truth mask (and vice-versa):

\begin{equation}
\begin{aligned}
T_{prec}(S(P), G) &= \frac{|S(P) \cap G|}{|S(P)|}, \\
T_{sens}(S(G), P) &= \frac{|S(G) \cap P|}{|S(G)|},
\end{aligned}
\end{equation}

The clDice index is the harmonic mean of these measures. The loss is minimized to ensure the network preserves the graph-like structure of the vasculature:

\begin{equation}
L_{clDice} = 1 - \frac{2 \, T_{prec} \, T_{sens}}{T_{prec} + T_{sens}},
\end{equation}

\textbf{Composite Objective: }The final objective function $L_{total}$ is a weighted sum of the region-based Tversky loss and the topology-aware clDice loss:

\begin{equation}
L_{total} = L_{Tv} + \lambda L_{clDice},
\end{equation}

Based on our empirical evaluations, we set the weighting factor $\lambda = 0.5$. This configuration balances volumetric segmentation accuracy with the preservation of vascular connectivity across both artery and vein channels.

\section{Experiments}
\label{sec:experiments}

\subsection{Dataset}
\label{subsec:dataset}
We use the \textbf{Fundus-AVSeg} dataset~\cite{deng2025fundus}, a recent benchmark for artery–vein segmentation containing $100$ high-resolution fundus images ($2656\times1992$ or $1280\times1280$) captured with ZEISS VISUCAM200 and Canon cameras at Shenzhen Eye Hospital. Annotations were produced through a multi-stage protocol involving six junior and one senior ophthalmologist, providing pixel-level labels for arteries, veins, crossings, and uncertain vessels.

\subsection{Implementation Details}
\label{subsec:implementation_details}

\textbf{Data Pre-processing and Augmentation:}
The dataset was randomly partitioned into training ($80\%$), validation ($10\%$), and testing ($10\%$) subsets. To ensure experimental reproducibility, a fixed random seed was utilized for the split. All input images and corresponding masks were resized to a spatial resolution of $512 \times 512$ pixels. Intensity scaling was applied to the images to normalize the pixel value range. We implemented a comprehensive online data augmentation pipeline using the MONAI library. Table \ref{tab:augmentations} shows the details of the augmentation strategies applied during the training phase.

\begin{table}[h]
\centering
\resizebox{\columnwidth}{!}{%
\begin{tabular}{llc}
    \toprule
    \textbf{Augmentation Technique} & \textbf{Parameters} & \textbf{Probability} \\
    \midrule
    Random Rotation & Range $\in [-30^{\circ}, +30^{\circ}]$ & 0.5 \\
    Vertical Flip & - & 0.5 \\
    Horizontal Flip & - & 0.5 \\
    Affine Translation & Shift limit $= 5\%$ & 0.3 \\
    Random Contrast & $\gamma \in [0.8, 1.2]$ & 0.3 \\
    Intensity Shift & Offset $= 0.1$ & 0.3 \\
    Gaussian Noise & $\mu=0, \sigma=0.01$ & 0.2 \\
    Gaussian Smoothing & $\sigma \in [0.5, 1.0]$ & 0.2 \\
    \bottomrule
\end{tabular}%
}
\caption{Data augmentation techniques with hyperparameters.}
\label{tab:augmentations}
\vspace{-.9em}
\end{table}

\textbf{TFFM Configuration:}
To effectively inject global structural awareness into the local receptive fields, we implemented the TFFM with scale-adaptive hyperparameters. As direct graph construction on $512^2$ pixels is computationally prohibitive, we utilized the module's adaptive pooling to transform feature maps into manageable coarse grids, thereby shifting the paradigm from spatial proximity to content-based similarity. The TFFM was applied at decoder levels $l \in \{0, 1, 2, 3, 4\}$ to resolve the tunnel vision limitation of standard convolutions. The grid sizes ($G_l$) were configured progressively as $\{20, 24, 28, 32, 32\}$, with corresponding neighbor counts ($k_l$) of $\{5, 7, 9, 12, 15\}$. This configuration ensures that deeper levels (with higher $k$) capture broader global dependencies to maintain vessel continuity, while shallower levels preserve local feature fidelity.

\textbf{Training Setup:}
The proposed framework was implemented in PyTorch and MONAI. All experiments were conducted on an NVIDIA H200 GPU. The model was optimized using the AdamW optimizer with an initial learning rate of $1 \times 10^{-3}$. We utilized a batch size of $10$. The training process was set to run for a maximum of $500$ epochs, incorporating an early stopping mechanism with a patience of $10$ epochs. Training was halted if the validation loss did not improve within this window.

\textbf{Experimental Progression:}
To isolate the contribution of each architectural component, we followed a systematic ablation strategy where only one variable was modified at a time. The overall workflow is illustrated in Figure \ref{fig:progression} and detailed below:

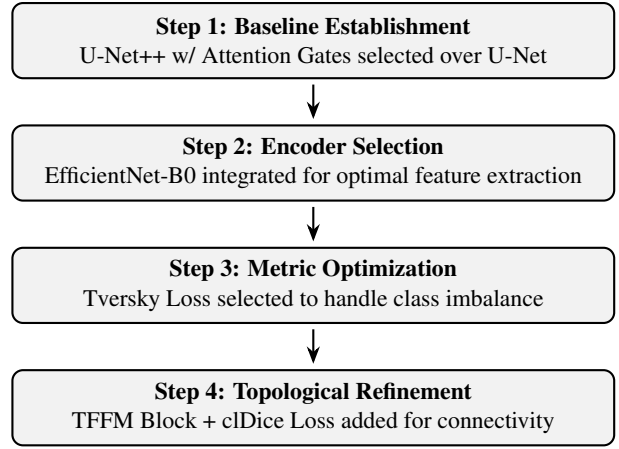
\begin{figure}[ht]
    \centering
    \begin{tikzpicture}[
        node distance=0.6cm,
        stepbox/.style={
            rectangle, 
            draw=black, 
            thick,
            fill=gray!10,
            align=center, 
            rounded corners, 
            minimum width=8cm, 
            minimum height=1.0cm,
            font=\small
        },
        arrow/.style={
            -Stealth, 
            thick,
            shorten >=2pt,
            shorten <=2pt
        }
    ]
        \node (step1) [stepbox] {\textbf{Step 1: Baseline Establishment} \\ U-Net++ w/ Attention Gates selected over U-Net};
        \node (step2) [stepbox, below=of step1] {\textbf{Step 2: Encoder Selection} \\ EfficientNet-B0 integrated for optimal feature extraction};
        \node (step3) [stepbox, below=of step2] {\textbf{Step 3: Metric Optimization} \\ Tversky Loss selected to handle class imbalance};
        \node (step4) [stepbox, below=of step3] {\textbf{Step 4: Topological Refinement} \\ TFFM Block + clDice Loss added for connectivity};
        \draw[arrow] (step1) -- (step2);
        \draw[arrow] (step2) -- (step3);
        \draw[arrow] (step3) -- (step4);
    \end{tikzpicture}
    \caption{Systematic experimental progression used to derive the final topology-aware framework.}
    \label{fig:progression}
\end{figure}

\subsection{Evaluation Metrics}
\label{subsec:metrics}
To fully assess the performance of our model, we employ two categories of metrics: pixel-wise segmentation metrics and topology-aware structural metrics.

\textbf{Pixel-wise Metrics:} We report the Dice Similarity Coefficient (DSC) and Intersection over Union (IoU) to measure region overlap. To evaluate boundary accuracy, which is critical for thin vessels, we compute the $95$\% Hausdorff Distance (HD95). HD95 is computed separately for arteries and veins against their respective ground truth, and on merged masks for combined vessel segmentation.

\textbf{Topological and Connectivity Metrics:} Standard pixel metrics often fail to penalize broken vessel connections. Therefore, we include:
\begin{itemize}
    \item \textbf{clDice~\cite{shit2021cldice}:} A topology-preserving metric that calculates the harmonic mean of topological precision and sensitivity based on the vascular skeleton.
    \item \textbf{Betti Number Error (Betti0-Err):} This metric measures the difference in the number of connected components between prediction and ground truth. A lower error indicates better preservation of vascular connectivity, with fewer fragmented vessel segments.
    \item \textbf{Skeleton-F1 and Junction-Recall:} We skeletonize the prediction and ground truth to measure the F1-score of the vessel centerlines and the recall rate of bifurcation points (junctions), which are essential for graph-based downstream analysis.
\end{itemize}

\section{Results}
\label{sec:results}

To establish a robust framework for topology-aware vessel segmentation, we adopted a systematic ablation strategy. We optimize the pipeline sequentially shown in Section \ref{subsec:implementation_details}.

\subsection{Baseline Architecture Selection}
\label{subsec:baseline_selection}

We first evaluated segmentation architectures under controlled conditions (BCE loss, AdamW, no augmentation, no pre-trained encoder). Table \ref{tab:baseline_comparison} presents the performance on the combined vasculature.

\begin{table}[H]
\centering
\resizebox{\columnwidth}{!}{%
\begin{tabular}{lccccc}
\toprule
\textbf{Architecture} & \textbf{IoU(\%)} & \textbf{Dice(\%)} & \textbf{HD95} & \textbf{Precision(\%)} & \textbf{Recall(\%)} \\
\midrule
U-Net & 62.86 & 77.10 & 12.45 & 94.11 & 65.67 \\
Attention U-Net & 81.45 & 89.75 & 4.60 & 93.40 & 86.55 \\
U-Net++ & 81.69 & 89.89 & 4.27 & \textbf{95.19} & 85.36 \\
\textbf{U-Net++ w/ Attn} & \textbf{83.52} & \textbf{91.00} & \textbf{3.59} & 94.05 & \textbf{88.31} \\
SegResNet & 79.08 & 88.29 & 7.05 & 94.05 & 83.36 \\
TransUNet (ViT-B) & 73.29 & 84.56 & 8.34 & 90.19 & 79.75 \\
SwinUNETR & 74.76 & 85.54 & 9.97 & 95.04 & 77.94 \\
DeepLabV3+ & 49.02 & 65.76 & 11.00 & 73.28 & 59.77 \\
Segformer & 66.00 & 79.48 & 7.78 & 85.19 & 74.72 \\
\bottomrule
\end{tabular}%
}
\caption{Baseline architectural comparison (combined).}
\label{tab:baseline_comparison}
\vspace{-1em}
\end{table}

As Unet++ and Attention U-Net performed well, we integrated U-Net++ with Attention Gates, which achieved the highest Dice ($91$\%) and lowest boundary error (HD95: $3.59$). The nested skip connections preserved multi-scale features, while Attention Gates successfully improved recall to $88.31$\% without sacrificing precision ($94.05$\%). Transformer-based models (TransUNet, SwinUNETR) underperformed in this data-constrained regime compared to the nested convolutional approach.

\subsection{Baseline Loss Function Optimization}
\label{subsec:loss_optimization}

To identify a loss function suitable for both class imbalance and topological continuity, we evaluated BCEDice, BoundaryDoU, LogCoshDice, and Tversky Loss ($\alpha=0.65$) using a fixed U-Net++ with Attention Gates architecture.

A qualitative summary of strengths and weaknesses is provided in Table \ref{tab:loss_comparison}. Quantitatively, LogCoshDice achieved the highest scores in pixel-level accuracy (Dice: $92.04\%$, HD95: $2.98$) and centerline metrics (Cent-Dice: $75.49\%$). Conversely, BoundaryDoU excelled in topological continuity, recording the lowest Betti0 error ($25.5$) and the least fragmentation ($27.2$ components, avg. length $310.20$). The Tversky loss provided a robust balance between these two profiles, consistently achieving the second-best result in nearly every category, including topology (Betti0 error: $27.3$), fragmentation, and centerline accuracy without placing first in pixel metrics. Given that vessel segmentation prioritizes topological fidelity, Tversky loss was selected as the optimal baseline for subsequent experiments.

\begin{table}[ht]
\centering\small
\begin{tabular}{p{1.8cm}p{2.5cm}p{2.5cm}}
\toprule
\textbf{Loss} & \textbf{Strengths} & \textbf{Weaknesses} \\
\midrule
BCEDice & Simple baseline with reasonable recall & Highest fragmentation and weak topology \\
BoundaryDoU & Lowest topological error and minimal fragmentation & Blurred boundaries and reduced connectivity \\
LogCoshDice & Best pixel metrics and highest Skel-F1 & Higher topological error and moderate fragmentation \\
\textbf{Tversky ($\alpha=0.65$)} & Balanced topology, connectivity, and boundary accuracy & Slightly below LogCoshDice in pixel metrics \\
\bottomrule
\end{tabular}
\caption{Qualitative comparison of the evaluated loss functions.}
\label{tab:loss_comparison}
\vspace{-1em}
\end{table}

\subsection{Encoder Selection}
\label{subsec:encoder_selection}

We evaluated five pretrained encoders integrated within the U-Net++ (Attn) + Tversky pipeline. Table~\ref{tab:encoder_comparison_updated} presents the comparative performance across key metrics.

\begin{table}[h]
\centering
\resizebox{\columnwidth}{!}{%
\begin{tabular}{lcccc}
\toprule
\textbf{Encoder} & \textbf{Metric} & \textbf{Artery} & \textbf{Vein} & \textbf{Combined} \\
\midrule
\textbf{EffNet-B0} & Dice (\%) $\uparrow$ & \textbf{81.80} & \textbf{84.32} & 90.15 \\
  & HD95 $\downarrow$ & \textbf{25.08} & \textbf{19.05} & \\
 & Betti0-Err $\downarrow$ & \textbf{38.60} & \textbf{38.00} & \\
\midrule
ResNet-101 & Dice (\%) $\uparrow$ & 80.98 & 84.05 & \textbf{90.63} \\
 & HD95 $\downarrow$ & 27.88 & 20.83 & \\
 & Betti0-Err $\downarrow$ & 44.40 & 34.80 & \\
\midrule
ConvNeXt-B & Dice (\%) $\uparrow$ & 80.96 & 83.98 & 90.24 \\
 & HD95 $\downarrow$ & 28.34 & 20.67 & \\
 & Betti0-Err $\downarrow$ & 48.10 & 44.40 & \\
\midrule
EffNet-B4 & Dice (\%) $\uparrow$ & 79.11 & 82.14 & 90.27 \\
 & HD95 $\downarrow$ & 26.94 & 23.00 & \\
 & Betti0-Err $\downarrow$ & 38.40 & 46.40 & \\
\midrule
ResNet-50 & Dice (\%) $\uparrow$ & 80.56 & 83.44 & 89.89 \\
 & HD95 $\downarrow$ & 27.33 & 19.67 & \\
 & Betti0-Err $\downarrow$ & 51.60 & 36.10 & \\
\bottomrule
\end{tabular}%
}
\caption{Encoder comparison (U-Net++ Attn + Tversky).}
\label{tab:encoder_comparison_updated}
\vspace{-1em}
\end{table}

EfficientNet-B0 delivered the best individual vessel performance, achieving the highest Dice scores for both artery ($81.80\%$) and vein ($84.32\%$) while substantially improving boundary accuracy (HD95: $25.08/19.05$). Notably, it outperformed the larger EfficientNet-B4 despite using $75$\% fewer parameters, suggesting superior model-dataset fit. While ResNet-101 achieved marginally higher combined-class Dice ($90.63\%$), it exhibited greater fragmentation in arteries (Betti0-Error: $44.40$ vs $38.60$) and inferior sensitivity to fine vessel details. EfficientNet-B0's balance of per-class accuracy, minimal topological fragmentation, and computational efficiency established it as the optimal encoder.

\subsection{Topological Refinement (TFFM)}
\label{subsec:tffm_results}

Having fixed the backbone (U-Net++ w/ Attn), loss (Tversky), and encoder (EfficientNet-B0), we introduced the proposed TFFM.

\begin{table}[H]
\centering
\resizebox{\columnwidth}{!}{%
\begin{tabular}{lcccc}
\toprule
\textbf{Configuration} & \textbf{Comb. Dice(\%)} & \textbf{clDice(\%)} & \textbf{Betti0-Err} & \textbf{Junc-F1(\%)} \\
\midrule
EffNet-B0 (Baseline) & 90.15 & 84.54 & 43.0 & 63.64 \\
\textbf{Baseline + TFFM} & \textbf{90.47} & \textbf{85.05} & \textbf{26.0} & \textbf{65.44} \\
\bottomrule
\end{tabular}%
}
\caption{Impact of TFFM on topology.}
\label{tab:tffm_impact}
\vspace{-1em}
\end{table}

As shown in Table~\ref{tab:tffm_impact}, TFFM integration improved topological fidelity: Betti number error dropped by $39.5$\% ($43.0$ →  $26.0$), and junction detection F1-score increased from $63.64\%$ to \textbf{$65.44$\%}. These gains were accompanied by modest improvements in per-class segmentation (Artery Dice: $81.80\%$ →  \textbf{$82.39\%$}, Vein Dice: $84.32\%$ →  \textbf{$84.77\%$}) and reduced Hausdorff distances. The module also reduced over-fragmentation: predicted components decreased from $44.7$ to $27.7$ ($38$\% reduction), aligning better with the ground truth topology.

\subsection{Final Model: Topological Loss and Augmentation}
\label{subsec:final_model}

To maximize generalization and connectivity, we incorporated the soft clDice loss ($L_{total} = L_{Tv} + 0.5 L_{clDice}$) into the TFFM framework and applied a comprehensive augmentation pipeline (realistic rotations, flips, translation, brightness/contrast adjustments, Gaussian noise/blur). This final configuration achieved a validation Dice of $87.20\%$ at epoch $72$, substantially improving generalization over the non-augmented TFFM model ($83.31\%$).

\begin{table}[H]
\centering
\resizebox{\columnwidth}{!}{%
\begin{tabular}{lccc}
\toprule
\textbf{Metric} & \textbf{Artery} & \textbf{Vein} & \textbf{Combined} \\
\midrule
Dice Score(\%) $\uparrow$ & 85.75 & 87.63 & 90.97 \\
HD95 (px) $\downarrow$ & 21.22 & 14.59 & 3.50 \\
clDice(\%) $\uparrow$ & 79.07 & 81.46 & 85.55 \\
Pred. Components $\downarrow$ & 23.3 & 26.1 & 25.3 \\
\bottomrule
\end{tabular}%
}
\caption{Final proposed framework performance.}
\label{tab:final_results}
\vspace{-1.1em}
\end{table}

The final model delivered improvements across all metrics (Table \ref{tab:final_results}). Artery and vein Dice scores increased to $85.75\%$ and $87.63\%$, respectively, while HD95 distances decreased to $21.22$ and $14.59$, enhancing boundary accuracy for thin vessels. The clDice metric improved substantially to $85.55\%$, with skeleton F1 scores reaching $87.63\%$ for arteries and $88.48\%$ for veins. Most critically, fragmentation was minimized to $25.3$ predicted components, the lowest achieved, yielding highly continuous vessel trees. Junction detection precision improved to $0.63 / 0.71$ for artery/vein, though recall remains limited at $ 0.44 / 0.52$, suggesting future work should focus on branch point completeness. These results confirm that the framework produces topologically robust segmentations suitable for automated graph analysis.

\section{Cross-Dataset Generalization and Transferability Analysis}
\label{sec:cross_dataset}
To rigorously evaluate the generalization capability of our proposed TFFM architecture, we conducted an extensive cross-dataset validation study. Unlike conventional within-dataset evaluations, we performed zero-shot inference on five established retinal vessel segmentation benchmarks without any fine-tuning or adaptation. This protocol assesses the model's robustness to variations in imaging protocols, scanner types, and pathological presentations.

\subsection{Experimental Protocol}
\label{subsec:cross_protocol}
The model trained on Fundus-AVSeg (Section \ref{subsec:dataset}) was directly applied to external test sets. All images were pre-processed identically, resized to $512\times512$ pixels, and intensity-normalized. No augmentation was applied during inference. We report comprehensive metrics that span pixel-wise accuracy, topological fidelity, and junction detection to provide a holistic assessment of domain adaptation.

\subsection{Cross-Dataset Quantitative Performance}
\label{subsec:cross_results}
Table \ref{tab:cross_dataset_summary} presents the aggregated performance metrics across five publicly available datasets representing diverse imaging conditions. Our framework maintained strong segmentation quality and, notably, preserved topological consistency.

\begin{table}[h]
\centering
\resizebox{\columnwidth}{!}{%
\begin{tabular}{lccccc}
\hline
\textbf{Dataset} & \textbf{Dice(\%)} ↑  & \textbf{HD95} ↓  & \textbf{clDice(\%)} ↑  & \textbf{Betti0-Err} ↓  & \textbf{Junc-F1(\%)} ↑  \\
\hline
DRIVE \cite{drive} & 82.10 & 10.20 & 70.67 & 24.20 & 30.63 \\
CHASEDB1 \cite{chasedb1} & 80.61 & 12.77 & 68.71 & 50.50 & 41.42 \\
HRF \cite{budai_robust_2013} & 79.94 & 12.67 & 73.01 & 33.00 & 55.10 \\
RETA \cite{reta2022} & 82.18 & 15.31 & 73.57 & 34.57 & 51.34 \\
STARE \cite{stare} & 80.70 & 33.25 & 70.98 & 41.55 & 43.34 \\
\hline
\end{tabular}%
}
\caption{Cross-dataset inference performance (combined vasculature).}
\label{tab:cross_dataset_summary}
\vspace{-.5cm}
\end{table}

\textbf{Vessel Segmentation Accuracy:} The model achieved Dice coefficients ranging from $79.94\%$ (HRF) to $82.18\%$ (RETA), showing robust pixel-level performance across domains. Notably, DRIVE and RETA results approached the performance on the source domain (Fundus-AVSeg: $82.18\%$ vs. $90.97\%$), indicating strong alignment with healthy retinal morphology. The elevated HD95 on STARE ($33.25$ px) and RETA ($15.31$ px) reflects challenges with pathological cases and high myopia, where vessel caliber and tortuosity differ significantly from training data.

\textbf{Topological Consistency:} The clDice metric remained consistently high ($68.71\%$--$73.57\%$), substantially outperforming traditional methods in zero-shot settings. The lowest Betti0-Error on DRIVE ($24.20$) confirms superior preservation of vascular connectivity for standard screening images. However, CHASEDB1 exhibited higher fragmentation (Betti0-Error: $50.50$), likely due to pediatric vasculature patterns unseen during training. Despite this, skeleton F1-scores remained strong ($82.68\%$--$87.41\%$), indicating that predicted centerlines remained spatially aligned even when minor breaks occurred.

\begin{figure}[ht]
    \centering
    \begin{subfigure}[b]{0.31\columnwidth}
        \includegraphics[width=\textwidth]{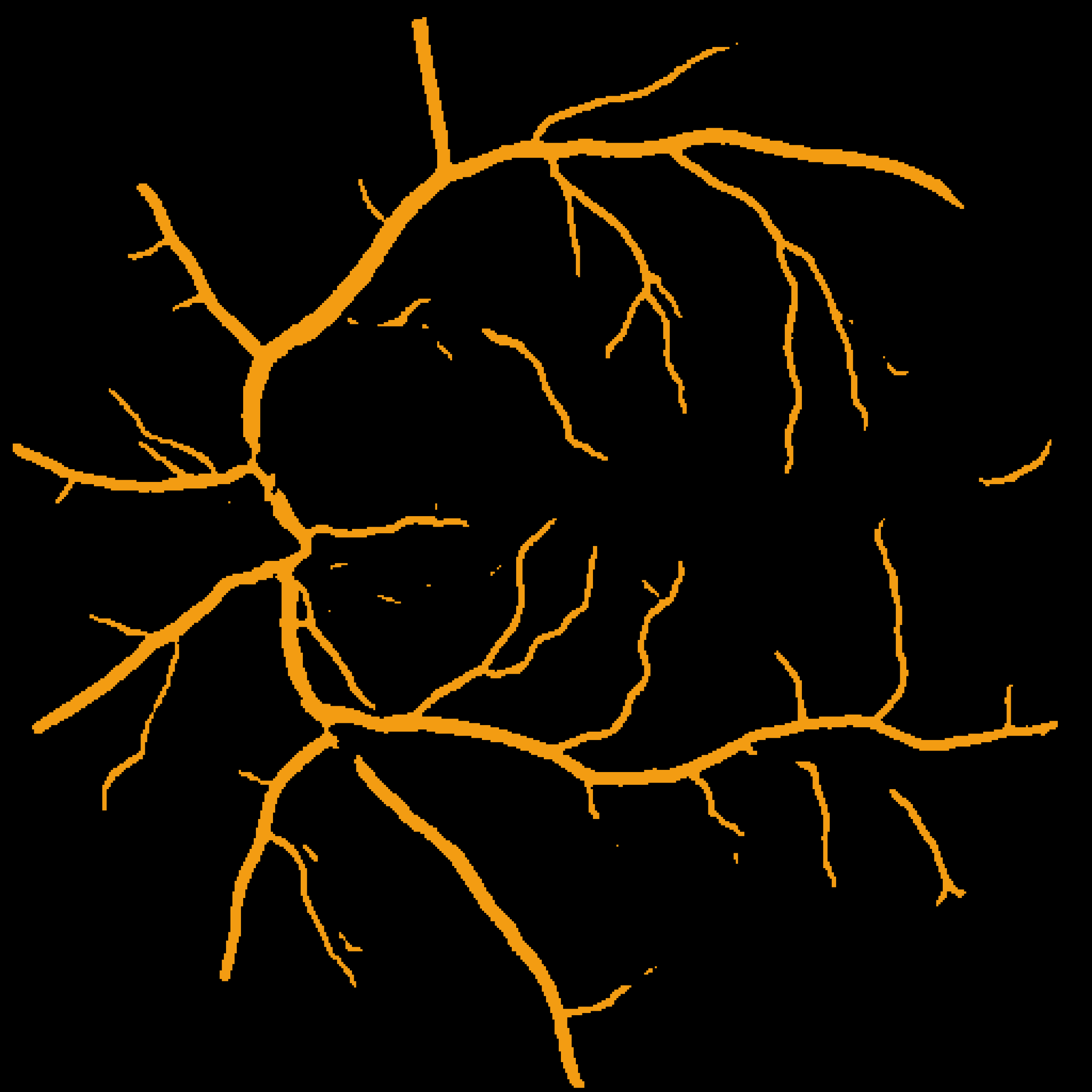}
        \caption{}
        \label{fig:sub1}
    \end{subfigure}
    \hfill
    \begin{subfigure}[b]{0.31\columnwidth}
        \includegraphics[width=\textwidth]{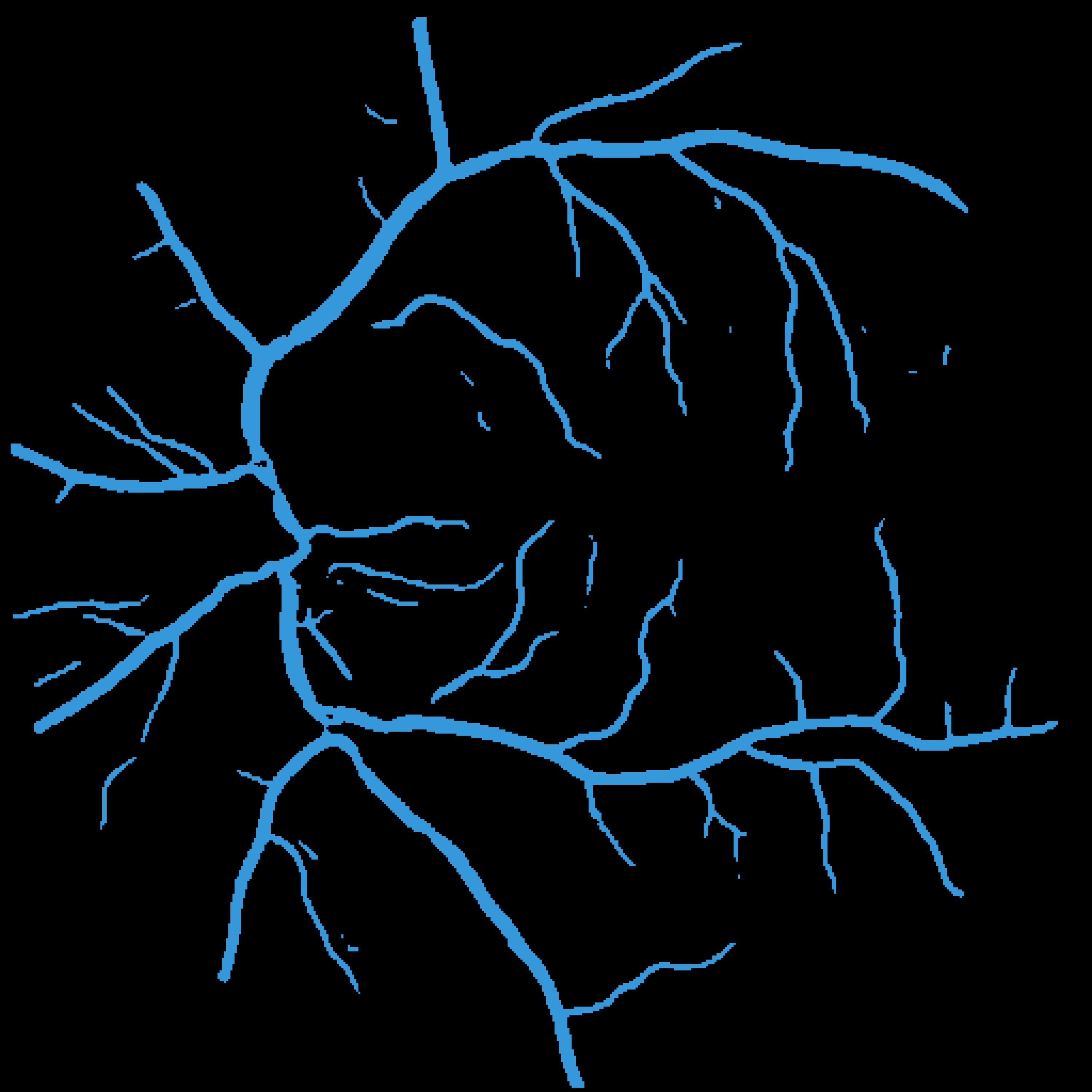}
        \caption{}
        \label{fig:sub2}
    \end{subfigure}
    \hfill
    \begin{subfigure}[b]{0.31\columnwidth}
        \includegraphics[width=\textwidth]{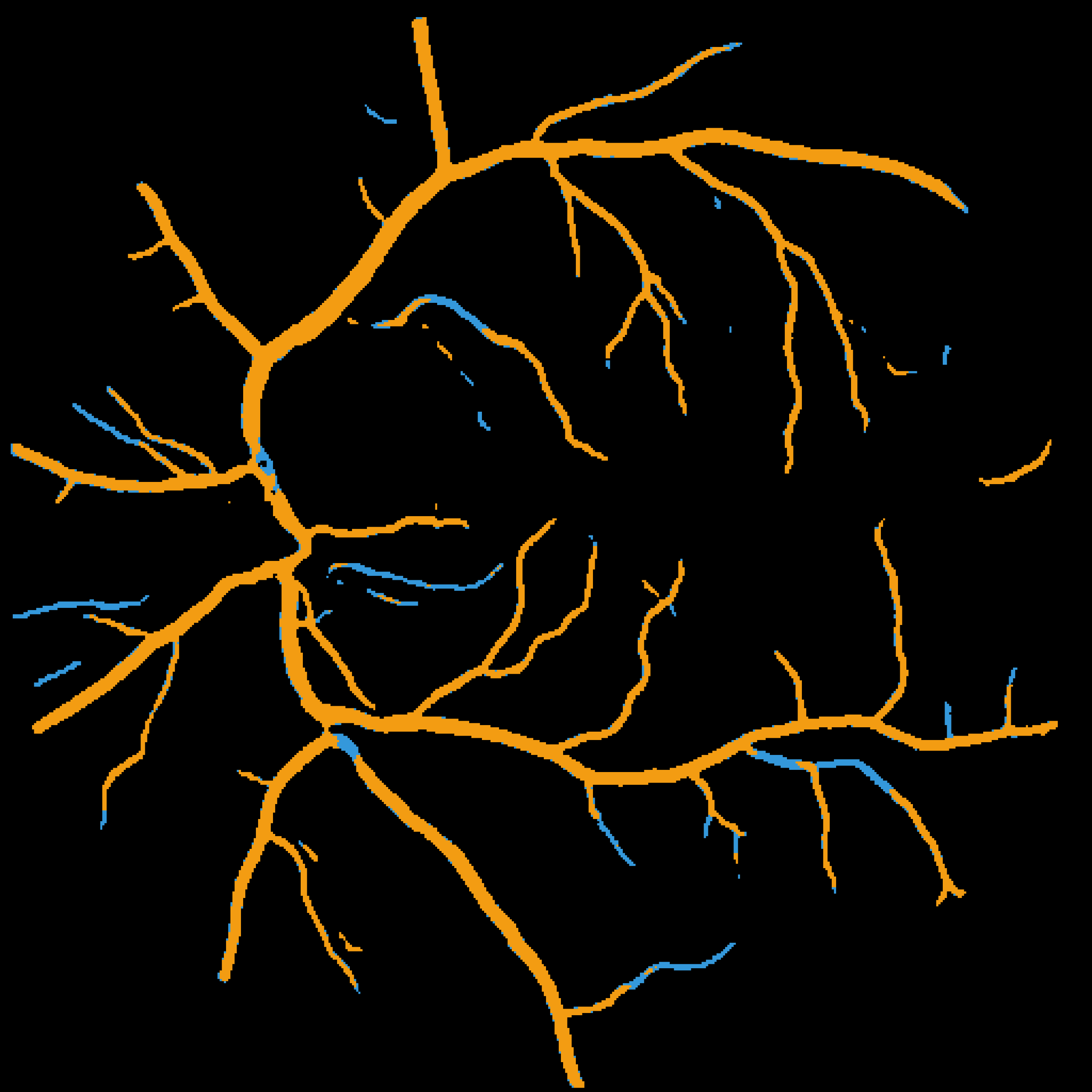}
        \caption{}
        \label{fig:sub3}
    \end{subfigure}
    
    \vspace{0.5em}
    
    \begin{subfigure}[b]{0.31\columnwidth}
        \includegraphics[width=\textwidth]{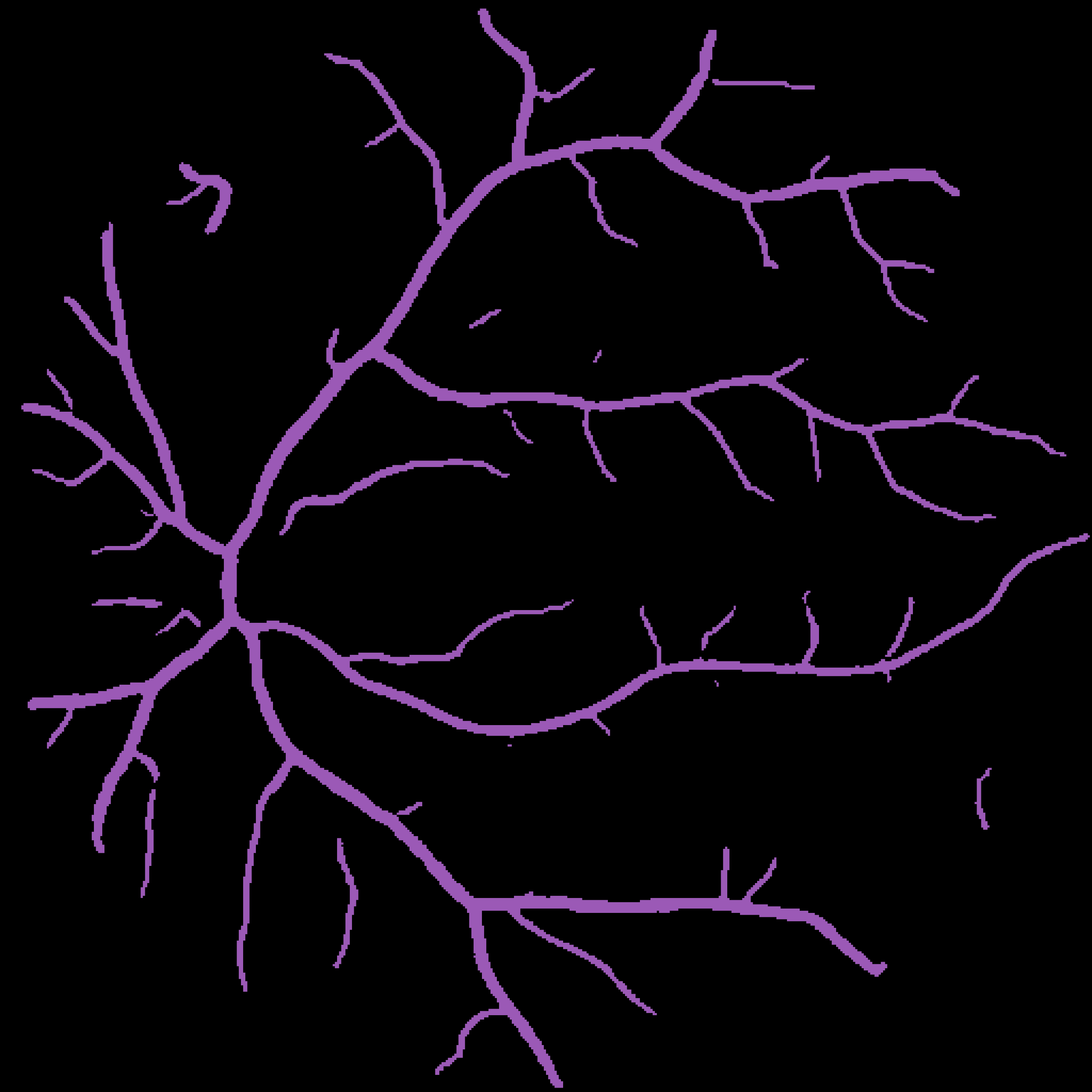}
        \caption{}
        \label{fig:sub4}
    \end{subfigure}
    \hfill
    \begin{subfigure}[b]{0.31\columnwidth}
        \includegraphics[width=\textwidth]{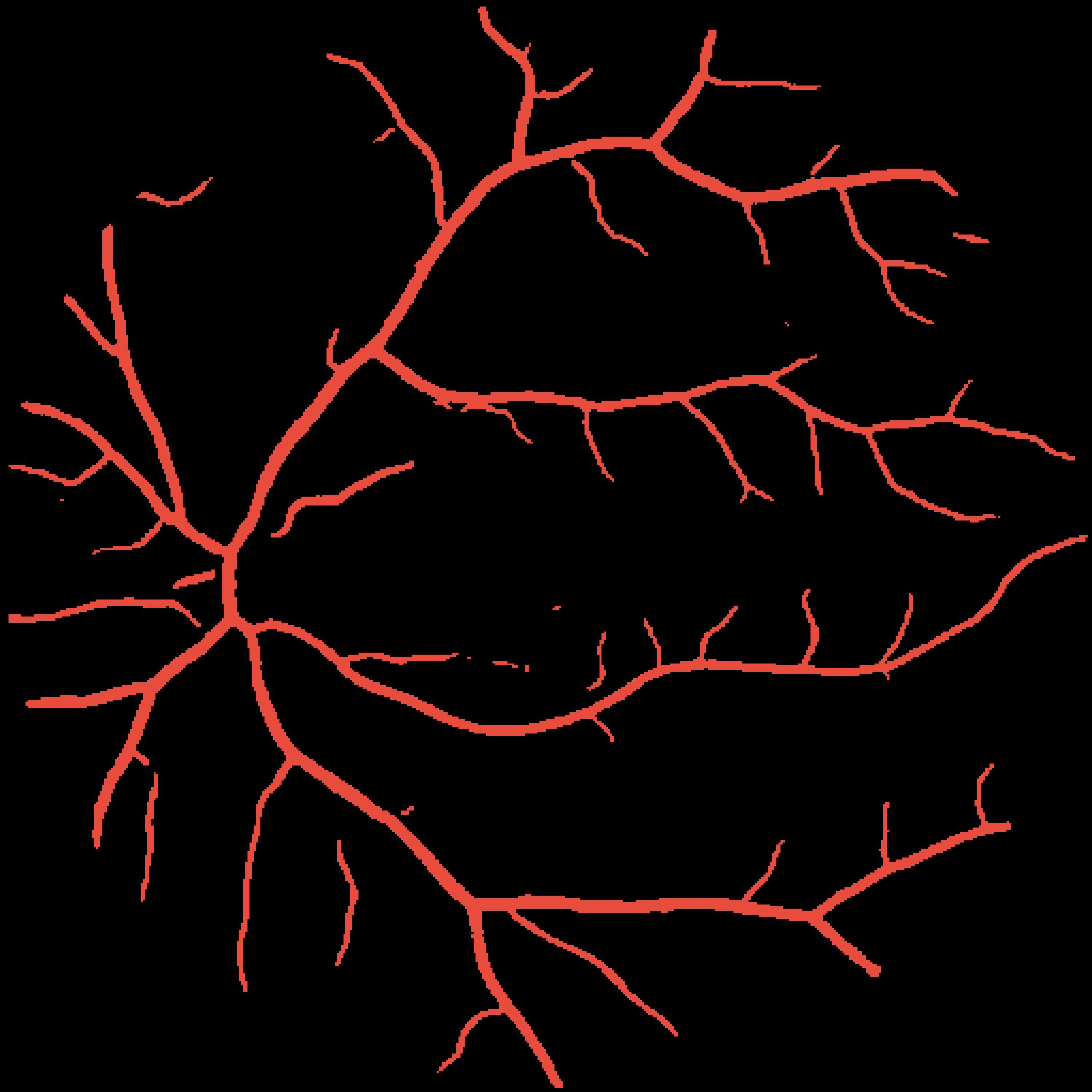}
        \caption{}
        \label{fig:sub5}
    \end{subfigure}
    \hfill
    \begin{subfigure}[b]{0.31\columnwidth}
        \includegraphics[width=\textwidth]{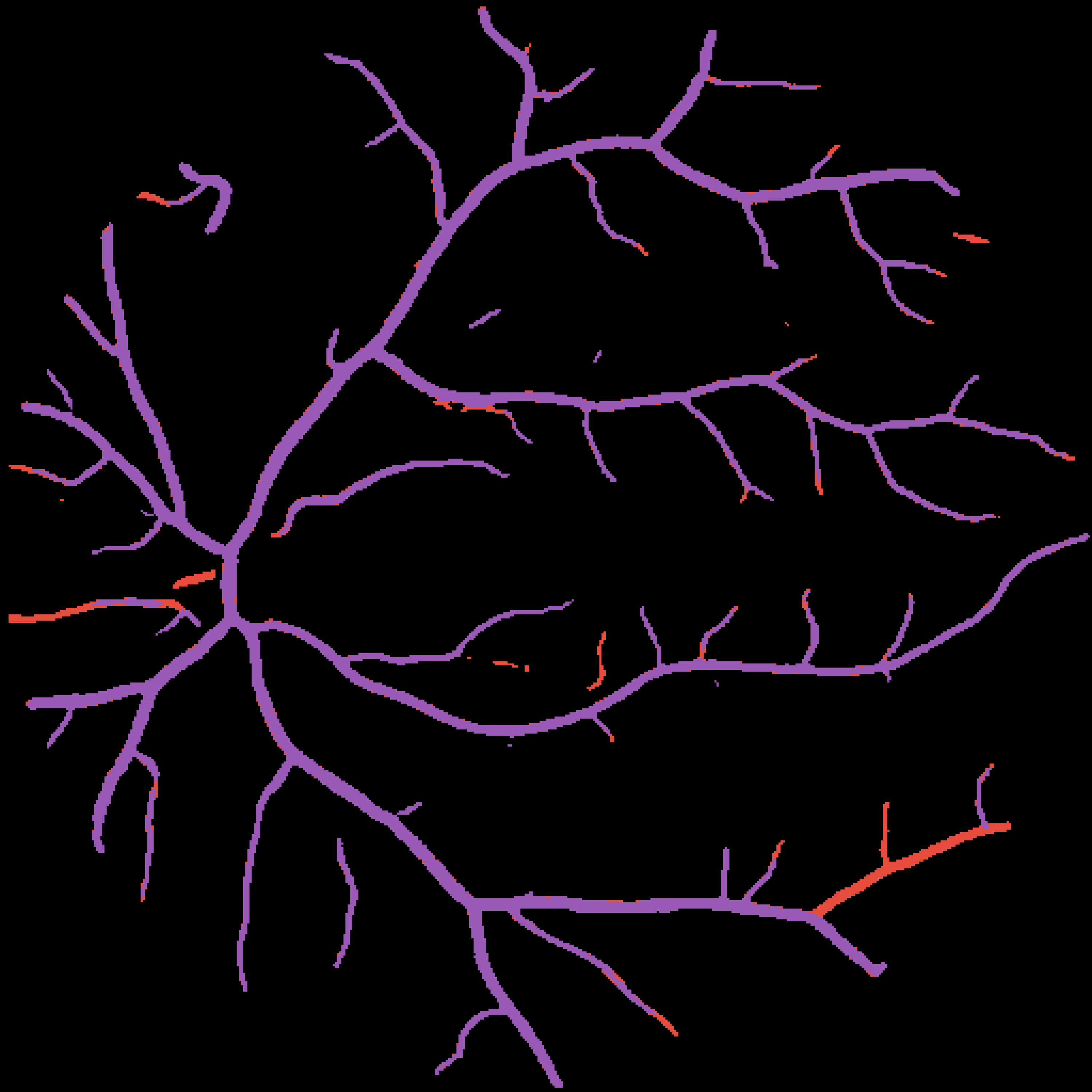}
        \caption{}
        \label{fig:sub6}
    \end{subfigure}
    \caption{Vein and Artery prediction visualization comparing performance with and without TFFM. Top row: (a) vein prediction without TFFM, (b) vein prediction with TFFM, (c) vein overlay. Bottom row: (d) artery prediction without TFFM, (e) artery prediction with TFFM, (f) artery overlay.}
    \label{fig:single_column_fig}
    \vspace{-1em}
\end{figure}

\section{Discussions}
\label{sec:discussion}

Our results highlight the critical disconnect between pixel-level accuracy and topological validity in retinal vessel segmentation. Baseline methods like LogCoshDice achieved high Dice scores ($92.04\%$) but failed to maintain structural integrity, with high topological error (Betti0-Err: $30.7$), corroborating the intrinsic topological vulnerability of standard CNNs that yield disjointed predictions clinically unusable for graph-based analysis. Integration of our TFFM resolved this disparity by mapping local features into a latent graph space, enabling the network to reason about global connectivity and reducing fragmented components by $38$\% (from $44.7$ to $27.7$) even before full augmentation, as visualized in Figure \ref{fig:single_column_fig}.

The ablation study validated our architectural choices: EfficientNet-B0 outperformed deeper encoders like ResNet-101, suggesting parameter-efficient networks better preserve fine spatial details for thin vessel segmentation. The relationship between Tversky loss (prioritizing recall) and soft clDice loss (penalizing topological breaks) was essential. TFFM integration alone reduced Betti error by $39.5$\%, a gain unachievable by pixel-based losses.  Crucially, this structural fidelity extends to zero-shot inference on external datasets, confirming that the learned topological constraints are robust to variations in scanner protocols and not merely overfitted to the training distribution. However, bifurcation completeness remains limited, with junction recall at $0.44$ for arteries and $0.52$ for veins, indicating the network still struggles with extreme local complexity at crossing points despite preserving global vessel tree connectivity, necessitating future specialized attention mechanisms. Clinically, these topological failures in complex bifurcations pose a risk for automated tortuosity quantification, which requires expert review for pathological cases. Future deployment on standard clinical workstations will require further validation to ensure these automated topological metrics correlate robustly with patient outcomes across diverse populations.

\section{Conclusions}
\label{sec:conclusion}

We introduced a topology-aware framework for retinal artery-vein segmentation that addresses vascular fragmentation by integrating TFFM with a hybrid objective function to capture global connectivity while maintaining local features. On the Fundus-AVSeg dataset, our method achieves state-of-the-art performance with $90.97\%$ combined Dice score, $3.50$ pixels HD95, and critically reduces fragmentation to $25.3$ components per image, significantly outperforming baselines. These results confirm our framework generates topologically robust segmentations advantageous for automated biomarker extraction.
Limited junction recall suggests bifurcation-aware attention mechanisms could improve branch point completeness. Future work includes extending to pathological datasets, lightweight variants for clinical deployment, and end-to-end graph architectures for arteriovenous classification and tortuosity quantification.

{
    \small
    \bibliographystyle{ieeenat_fullname}
    \bibliography{main}
}

\end{document}

%% file: preamble.tex
\usepackage{booktabs}
\usepackage{graphicx}
\usepackage{float}